\documentclass[letterpaper]{article} % DO NOT CHANGE THIS
\usepackage{aaai25}  % DO NOT CHANGE THIS
\usepackage{times}  % DO NOT CHANGE THIS
\usepackage{helvet}  % DO NOT CHANGE THIS
\usepackage{courier}  % DO NOT CHANGE THIS
\usepackage[hyphens]{url}  % DO NOT CHANGE THIS
\usepackage{graphicx} % DO NOT CHANGE THIS
\usepackage{natbib}  % DO NOT CHANGE THIS AND DO NOT ADD ANY OPTIONS TO IT
\usepackage{caption} % DO NOT CHANGE THIS AND DO NOT ADD ANY OPTIONS TO IT
\usepackage{algorithm}
\usepackage{algorithmic}

\usepackage{newfloat}
\usepackage{listings}
\DeclareCaptionStyle{ruled}{labelfont=normalfont,labelsep=colon,strut=off} % DO NOT CHANGE THIS
\floatstyle{ruled}
\newfloat{listing}{tb}{lst}{}
\floatname{listing}{Listing}
\usepackage{graphicx}
\usepackage{lscape}                                         % Useful for wide tables or figures.
\usepackage{threeparttable}

\usepackage{animate} %% convert frames to video

\usepackage{booktabs}                   % Publication quality tables
\usepackage{multirow}

\usepackage{makecell}

\usepackage{paralist}
\usepackage{enumitem}

\usepackage{epsfig}                      % for figures
\usepackage{graphicx}                  % another 
\usepackage{mathptmx}
\usepackage{mathtools}

\usepackage{url}  % Hyphenation of URLs.
\usepackage[table]{xcolor}
\usepackage{grfext}
\PrependGraphicsExtensions*{.jpg,.png,.PNG}

\usepackage{amssymb}% http://ctan.org/pkg/amssymb
\usepackage{pifont}% http://ctan.org/pkg/pifont
\def\eg{\textit{e.g.},~}               % for example
\def\ie{\textit{i.e.},~}               % that is, in other words
\def\etc{\textit{etc}}                 % and other things, and so forth
\def\vs{\textit{vs.}~}                 % against

\newcommand{\figref}[1]{Figure~\ref{fig:#1}} 

\newcommand{\appenref}[1]{Appendix \S ~\ref{appen:#1}}

\long\def\ignorethis#1{}

\newbox\jsavebox%

\def\xi{\mathbf{x}_i}

\graphicspath{{figure}, {example}}
\DeclareFixedFont{\ttb}{T1}{txtt}{bx}{n}{6} % for bold
\DeclareFixedFont{\ttm}{T1}{txtt}{m}{n}{6}  % for normal

\definecolor{codegreen}{rgb}{0,0.6,0}
\definecolor{codegray}{rgb}{0.5,0.5,0.5}
\definecolor{codepurple}{rgb}{0.58,0,0.82}
\definecolor{backcolour}{rgb}{0.95,0.95,0.92}

\lstdefinestyle{mystyle}{
    commentstyle=\color{codegreen},
    keywordstyle=\color{blue},
    numberstyle=\tiny\color{codegray},
    stringstyle=\color{codepurple},
    basicstyle=\ttfamily\footnotesize,
    breakatwhitespace=false,         
    breaklines=true,                 
    captionpos=b,                    
    keepspaces=true,                 
    numbers=left,                    
    numbersep=5pt,                  
    showspaces=false,                
    showstringspaces=false,
    showtabs=false,                  
    tabsize=2
}

\title{Building a Multi-modal Spatiotemporal Expert for Zero-shot Action Recognition with CLIP}
\author{
    Yating Yu\equalcontrib,
    Congqi Cao\equalcontrib\thanks{Corresponding author.},
    Yueran Zhang,
    Qinyi Lv,
    Lingtong Min,
    Yanning Zhang
}
\affiliations{
    Northwestern Polytechnical University, Xi'an Shaanxi, 710129, China\\
    yatingyu@mail.nwpu.edu.cn, congqi.cao@nwpu.edu.cn, zhangyueran3@mail.nwpu.edu.cn, \{lvqinyi, minlingtong, ynzhang\}@nwpu.edu.cn
}

\begin{document}

\maketitle

\begin{abstract}
Zero-shot action recognition (ZSAR) requires \textit{collaborative multi-modal spatiotemporal understanding}. 
However, finetuning CLIP directly for ZSAR yields suboptimal performance, given its inherent constraints in capturing essential temporal dynamics from both vision and text perspectives, especially when encountering novel actions with fine-grained spatiotemporal discrepancies.
In this work, we propose \textbf{Spatiotemporal Dynamic Duo} (STDD), a novel CLIP-based framework to comprehend multi-modal spatiotemporal dynamics synergistically. 
For the vision side, we propose an efficient Space-time Cross Attention, which captures spatiotemporal dynamics flexibly with simple yet effective operations applied before and after spatial attention, without adding additional parameters or increasing computational complexity.
For the semantic side, we conduct spatiotemporal text augmentation by comprehensively constructing an Action Semantic Knowledge Graph (ASKG) to derive nuanced text prompts. The ASKG elaborates on static and dynamic concepts and their interrelations, based on the idea of decomposing actions into spatial appearances and temporal motions.
During the training phase, the frame-level video representations are meticulously aligned with prompt-level nuanced text representations, which are concurrently regulated by the video representations from the frozen CLIP to enhance generalizability. 
Extensive experiments validate the effectiveness of our approach, which consistently surpasses state-of-the-art approaches on popular video benchmarks (\ie Kinetics-600, UCF101, and HMDB51) under challenging ZSAR settings. 
% Code is available at \url{https://github.com/Mia-YatingYu/STDD}.
\end{abstract}

% Uncomment the following to link to your code, datasets, an extended version or similar.
%
\begin{links}
    \link{Code}{https://github.com/Mia-YatingYu/STDD}
    % \link{Datasets}{https://aaai.org/example/datasets}
    \link{Extended version}{https://arxiv.org/abs/2412.09895}
\end{links}

\section{Introduction}
\label{sec:intro}

\begin{figure}[t]
    \centering
    \includegraphics[width=\columnwidth]{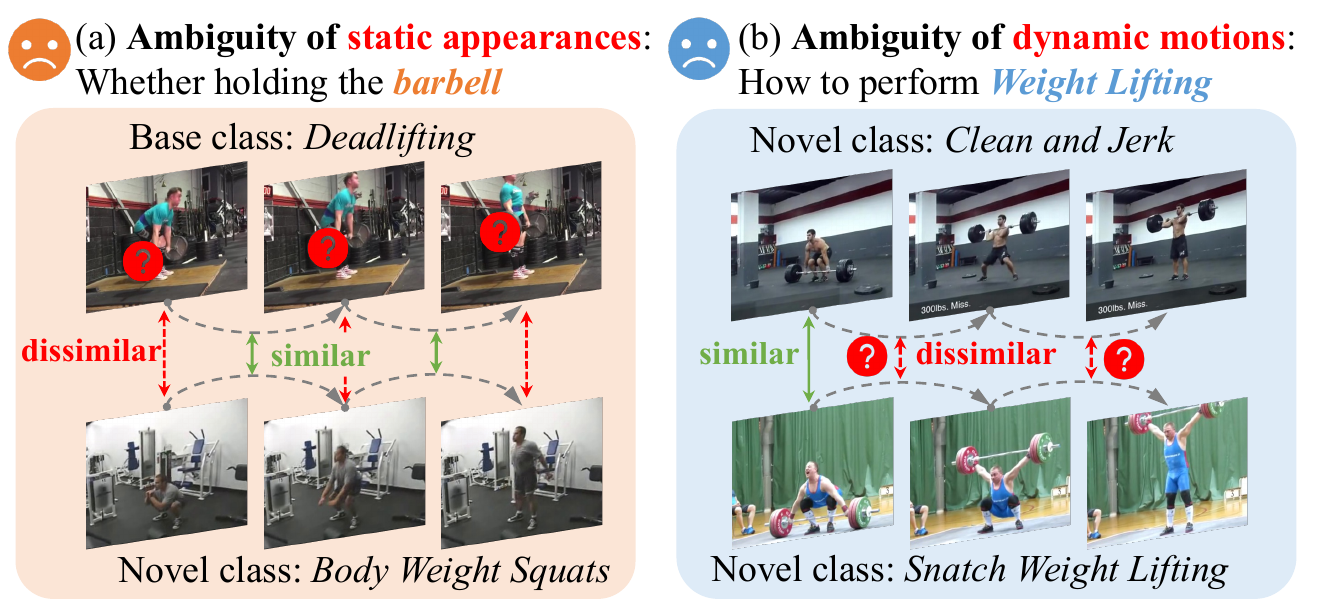}
    \caption{Illustration of the challenges without \textit{collaborative multi-modal spatiotemporal understanding}. (a) A model lacking static context alignment may misidentify the novel class due to the ambiguity associated with the \textit{barbell}. (b) It might also struggle generalizing to other novel weightlifting actions, due to the subtle dynamic differences and strong visual similarities.}
    \label{fig:teaser}
\end{figure}

% ZSAR:
Zero-shot action recognition (ZSAR) aims to classify video actions from novel categories that are not present in the training of models. 
A strong ZSAR learner should be endowed with \textit{collaborative multi-modal spatiotemporal understanding}, where the statics and dynamics of videos and semantics should be aligned meticulously.
Otherwise, it would inevitably lead to an ambiguous comprehension of actions.
Let's first delve into a simple example. In \figref{teaser}(a), a model lacking the ability to align visual contexts with static concepts \eg the \textit{barbell}, may confuse whether the actor is performing \textit{Dead Lifting} or just doing \textit{Body Weight Squats}. Conversely, as shown in \figref{teaser}(b), a model might struggle to generalize to novel actions of \textit{Clean and Jerk} and \textit{Snatch Weight Lifting}, if it falters in aligning nuanced multi-modal spatiotemporal dynamics of \textit{Weight Lifting}, as they exhibit significant static visual affinities. 
%%%%%%%% Figure1 %%%%%%%%%%%%%%%%
%%%%%%%% Figure2  %%%%%%%%%%%%%%%%
\begin{figure}[t]
    \centering
\includegraphics[width=\columnwidth]{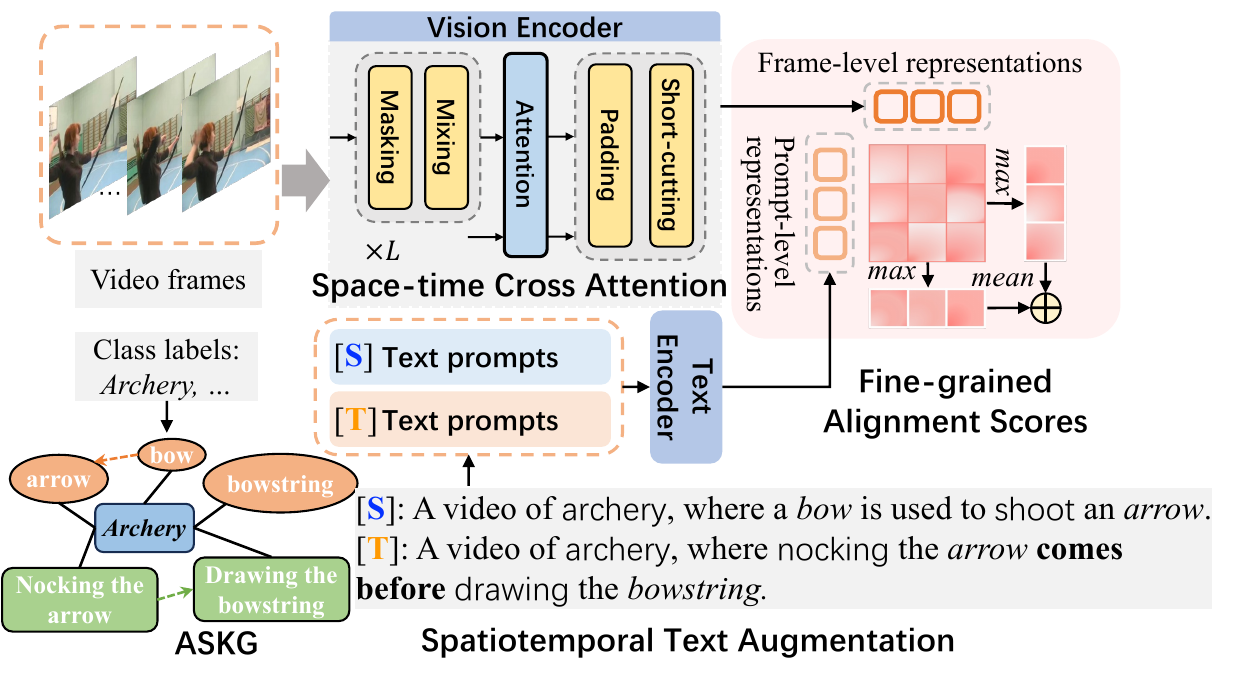}
    \caption{\textbf{Overview of our framework.} With a four-step operation applied within each block, we transform the spatial attention into novel Space-time Cross Attention. Spatiotemporal text augmentation is conducted to derive spatial and temporal text prompts, where multi-modal dynamics are meticulously aligned in a fine-grained manner.
    % integrates multi-modal spatiotemporal prompts and fine-grained alignment to serve as a ZSAR expert. We propose text augmentation for various text prompts from spatial statics [\textcolor{blue}{\textbf{S}}] and temporal dynamics [\textcolor{orange}{\textbf{T}}]. 
    }
    \label{fig:overview} 
\end{figure}

% existing works: temporal modeling vs semantic augmentaion
Contrastive Language-Image Pretraining (CLIP)~\cite{radford2021learning} has shown exceptional zero-shot inference in image-based tasks, benefiting from its strong generalization capability of the visual and linguistic alignment. Inspired by its success, CLIP can serve as a spatial expert for aligning static visual-semantic context of actions by processing the video frame-by-frame. However, due to its limitations in capturing temporal dynamics effectively, recent attempts~\cite{wu2023revisiting,wang2021actionclip,yang2023aim} have been made to adapt CLIP for general action recognition. Despite notable advancements obtained by additional temporal modeling~\cite{lin2022frozen,tu2023implicit,pan2022st}, they compromise with less-informative class-level prompts such as ``\textit{a video of} \{\verb|archery|\}'', thus faltering when encountering novel actions with fine-grained spatiotemporal discrepancies. 
Other works~\cite{chen2024ost,jia2024generating,wu2023bike} leverage Large Language Models (LLMs)~\cite{brown2020language,achiam2023gpt,floridi2020gpt} to extend CLIP with specialized knowledge, facilitating zero-shot generality. However, they often lay particular emphasis on semantic augmentation yet lacking efficient spatiotemporal understanding in visual world.

With these in mind, our key insight rests in extending the spatial expert \ie CLIP, into an effective spatiotemporal expert for ZSAR to comprehend multi-modal spatiotemporal dynamics synergistically.
To this end, we propose a novel CLIP-based framework, termed \textbf{Spatiotemporal Dynamic Duo} (STDD), which meticulously aligns spatiotemporal visual contexts with refined static and dynamic text prompts, as shown in~\figref{overview}.
Our framework enables flexible capture of spatiotemporal dynamics for efficient multi-scale cross-frame interaction without requiring additional parameters or increasing computational complexity.
Specifically, we realize this by implementing a four-step process \ie masking, mixing, padding and short-cutting, applied either before or after spatial attention. 
The masking strategy~\cite{qing2023mar} discards a proportion of visual tokens, and the subsequent channel mixing~\cite{bulat2021space} is performed with visible tokens at the interleaved positions across frames.
Then, the padding operation restores the original token quantity and spatial positions, while the ensuing short-cutting technique~\cite{he2016deep} is employed to seamlessly integrate dynamics into the primary vision encoding pathway.
The tailored combination of these operations jointly transforms the spatial attention into a novel Space-Time Cross Attention (STCA).
Besides, to acquire specialized action semantics systematically, we prompt LLMs to construct an Action Semantic Knowledge Graph (ASKG) that conceptually incorporates static appearances and dynamic motions of actions with a more structured and interpretable representation. 
Then, we perform spatiotemporal text augmentation to derive nuanced text prompts by parsing static and dynamic semantics in the ASKG.
Finally, frame-level video representations are meticulously aligned with prompt-level text representations, which are concurrently regulated by video representations from the frozen CLIP to enhance generalizability. 

Overall, our contributions can be summarized as follows:
\begin{itemize}
\item We introduce a novel CLIP-based framework, named \textbf{Spatiotemporal Dynamic Duo} (STDD), which synergistically comprehends dynamics for vision-text context refinement, facilitating \textit{multi-modal spatiotemporal understanding}. 

\item For the vision side, we propose to transform spatial attention into Space-time Cross Attention through a four-step operation to capture cross-frame dynamics without additional parameters or increased computational complexity. 

\item For the semantic side, we propose to perform spatiotemporal text augmentation by comprehensively constructing an Action Semantic Knowledge Graph, which articulates static appearances and dynamic motions of actions. 

\item Extensive experiments on three popular benchmarks verify the effectiveness and superiority of our method, consistently achieving state-of-the-art performance.

\end{itemize}

\section{Related Work}
\label{sec:related}
\paragraph{Adapting CLIP for Action Recognition.} 
% 
% Early works mainly harness visual features from off-the-shelf pretrained action recognition models and focus on modeling visual and semantic associations~\cite{brattoli2020rethinking,tran2018closer,gao2019know,ghosh2020all,mishra2020zero,mishra2018generative,wang2017alternative,wang2017zero,zhu2018towards,mandal2019out,zhang2018visual}. 

% 
Recently, Vision Language Models (VLMs)~\cite{sanghi2022clip,bao2022vlmo,yu2022coca} have demonstrated efficient multi-modal alignment, achieving impressive results in zero-shot inference. 
There is also a plethora of work~\cite{mao2024clip4hoi,qian2024intra,li2024zero} using knowledge learned in VLMs to video understanding tasks in a zero-shot manner. 
Adapting CLIP to videos~\cite{wang2021actionclip,wang2023seeing} is a common practice when designing a generalized video learner, where the key lies in utilizing additional temporal context for efficient video understanding. 
Recently, despite some works~\cite{wu2023revisiting,zhu2023orthogonal,wang2021actionclip} fully finetune the backbone with a video-header on top of CLIP, a collection of methods ~\cite{yang2023aim,lin2022frozen} work on parameter-efficient finetuning (PEFT)~\cite{gao2022visual,jie2022convolutional}, aiming to reduce trainable parameters, such as adapter-based~\cite{yang2023aim,lee2024cast,cao2024task}, prompt-based~\cite{wasim2023vita,ahmad2023ez} and decoder-based~\cite{lin2022frozen} methods.
There are also other works~\cite{huang2024froster,rasheed2023fine} exclude temporal modeling, while aiming to adapt features from image to videos by distilling knowledge from pretrained CLIP.
In contrast, our proposed framework facilitates temporal dynamic modeling without additional parameters, where the refined spatiotemporal representations are regulated by video representations from the frozen CLIP to preserve generalizability.

\paragraph{Space-time Self-Attention.}
% Historically, the remarkable success in the image domain frequently exerts a profound influence and shapes the architecture designs for the video domain. 
% 
Recently, the self-attention mechanism inherent in the ViT architecture~\cite{dosovitskiy2020image} for spatial modeling has been extensively adopted for video recognition. 
Due to the heavy complexity burden of full space-time attention, some prior works~\cite{bertasius2021space,arnab2021vivit} focus on factorizing spatial and temporal attention to adapt 3D data. 
AIM~\cite{yang2023aim} follows this idea by simply reusing pre-trained CLIP self-attention to perform temporal adaption, yet it nearly doubles the depth of the pre-trained encoder.
Open-VCLIP~\cite{weng2023open} expands the temporal attention view \ie dimension, for aggregating the global temporal information, maintaining the weight dimensions of the CLIP. 
Other variants~\cite{lin2019tsm,wang2022beyond} adopt the ``shift trick''~\cite{wu2018shift} with zero-cost dimensionality reduction to achieve temporal modeling at a layer level.
Most related to ours is X-ViT~\cite{bulat2021space}, which constructs the key vectors by mixing information from tokens located at the same spatial location within a local temporal window. 
We share the similar ``shift trick'', but our approach applies masks to spatial tokens ahead of mixing, which enables the mixing of information from tokens at interleaved spatial locations within a local spatiotemporal window.

\paragraph{Semantic Knowledge for Video-text Alignment.}
Semantic knowledge provides a bridge among actions, allowing the model to generalize to novel categories based on their semantic connections.
% % 
Early works usually design hand-crafted attributes~\cite{mandal2019out,mishra2020zero} or utilize object features~\cite{jain2015objects2action} to represent action semantics. 
% % 
Later, word-embedding methods~\cite{chen2021elaborative,wang2023visual,cao2024scene} are adopted for semantic representations. 
Recently, due to the versatility of LLMs, some works~\cite{chen2024ost,jia2024generating} construct knowledge-rich text descriptions by harnessing the responses from LLMs. 
And others~\cite{lin2023match,wu2024building} introduce the captions generated by BLIP~\cite{li2022blip,li2023blip} for multi-modal semantic knowledge.
% % 
% 
Diverging from existing semantic augmentation, we comprehensively construct a structured Action Semantic Knowledge Graph (ASKG) featuring refined static and dynamic semantics to derive spatial and temporal text prompts. Therefore, our spatiotemporal text augmentation allows for a more holistic representation of actions. 
% The meticulous alignment via video-text dynamic contexts significantly boosts collaborative multi-modal spatiotemporal understanding.

%%%%%%%%% Figure3  %%%%%%%%%%%%%%%%
\begin{figure*}[t]
    \centering
\includegraphics[width=\textwidth]{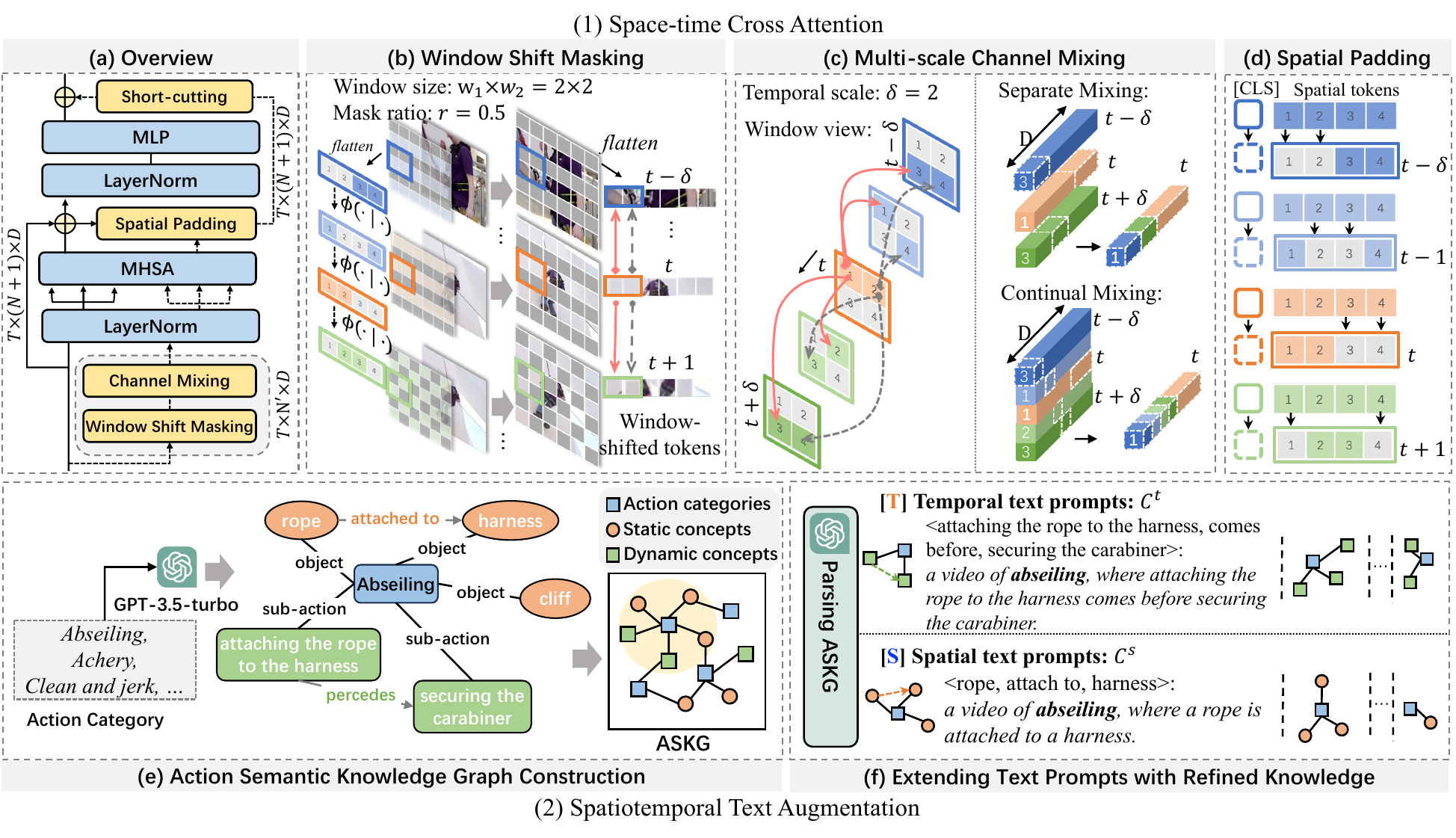}
    \caption{\textbf{Illustration of our method.} (1) We extend the spatial attention block to perform Space-time Cross Attention by applying Window Shift Masking to the input spatial tokens, and perform Multi-Scale Channel Mixing to capture temporal dynamics before MHSA. Then, we employ the spatial padding strategy to fill in the masked positions for seamless short-cutting, fusing additional dynamics effortlessly.  (2) We conduct spatiotemporal text augmentation to obtain nuanced spatial and temporal text prompts by elaborating on static and dynamic concepts with their interrelations presented in ASKG. }
    \label{fig:pipeline} 
\end{figure*}
\section{Method: Spatiotemporal Dynamic Duo}
\label{sec:method}
% \subsection{Problem Definition}
% \label{sec:problem}
% 
% ZSAR aims to identify unseen (or novel) action categories that are excluded from the training set by transferring knowledge from the seen domain $\mathcal{D}^s$ to unseen domain $\mathcal{D}^u$. The model is trained on videos from $\mathcal{D}^s=\{(V,y)\mid V\in V^s,y\in \mathcal{Y}^s\} $, where $V$ refers to a video clip, $y$ denotes its corresponding class label. Additionally, $\mathcal{D}^u=\{(V,y)\mid V\in V^u,y\in \mathcal{Y}^u\} $ is not available during training, whereas unseen category label set $\mathcal{Y}^u$ is available. In this setting, the category space is disjoint between $\mathcal{D}^s$ and $\mathcal{D}^u$, \ie $\mathcal{Y}^s \cap \mathcal{Y}^u=\emptyset $.

\subsection{Pipeline Overview}
\label{sec:pipeline}
Generally, as shown in \figref{overview}, our framework is capable of synergistic multi-modal spatiotemporal understanding which comprehends dynamics for vision-text context refinement.
Our model is initialized from the CLIP, which consists of a vision encoder $E_V$ and a text encoder $E_T$.

Specifically, given a video clip $V=\{\mathbf{x}_t\}_{t=1}^T,\mathbf{x}_t\in \mathbb{R}^{H\times W\times 3}$ of $T$ frames, by dividing each frame into $N$ non-overlapping patches $\{\mathbf{x}_{t,i}\}_{i=1}^N$ with the spatial size of $P\times P$, where $N=HW/P^2$. With prepending an additional [CLS] token $\mathbf{z}_{t,cls}^0$, to each frame, and adding positional embeddings to each patch, we can then obtain $\mathbf{z}_{t}^0 = [\mathbf{z}_{t,cls}^0,\mathbf{z}_{t,1}^0,\dots,\mathbf{z}_{t,N}^0] \in \mathbb{R}^{(N+1)\times D}$ via the patch embedding layer.

The $l$-th ViT block with our tailored Space-time Cross Attention consists of a Multi-head Self-Attention (\verb|MHSA|) followed by a \verb|MLP| layer with short-cutting (\verb|+|) and Layer Normalization (\verb|LN|). It processes the tokens $\mathbf{z}_{t}^{l-1}$ from the previous block as follows:
\begin{equation}
\label{eq:z}
    \mathbf{z'}_t^{l} = \verb|MHSA|(\verb|LN|(\mathbf{z}_t^{l-1}))+\mathbf{z}_t^{l-1},
\end{equation}
\begin{equation}
    \mathbf{z}_t^{l} = \verb|MLP|(\verb|LN|(\mathbf{z'}_t^{l}))+\mathbf{\tilde{z}}_t^l,
\end{equation}
where $\mathbf{\tilde{z}}_t^l$ is calculated by applying mixing-with-masking and padding before and after \verb|MHSA|, respectively, which will be introduced in detail in the next subsection. 
% Then, the short-cutting seamlessly assimilate into the encoding stream. 
Finally, the learned $\mathbf{z}_{t,cls}^L$ of the $t$-th frame from the last block is used as the frame-specific video representation.

Regarding the text flow, we perform \textit{spatial} and \textit{temporal} text augmentation for refined text prompts $C^{st}$ consisting of $C^s$ and $C^t$ which articulate the static appearances and dynamic motions of actions, respectively.
The $j$-th prompt-specific text representation $\mathbf{c}^{st}_{k,j}$ of the $k$-th class is obtained with the frozen $E_T$.

During training, we only optimize the parameters of $E_V$ by calculating the fine-grained alignment scores between $\mathbf{z}_{t,cls}^L$ and $\mathbf{c}^{st}_{k,j}$. Meanwhile, the video representation learning are regulated by the video representations from the pre-trained CLIP to distill knowledge.

\subsection{Space-time Cross Attention}
\label{sec:stca}
 As shown in~\figref{pipeline}(a), our proposed Space-time Cross Attention is primarily achieved by implementing a four-step process within each ViT block. Before \verb|MHSA|, (1) the Window Shift Masking (WSM) operation masks visual tokens and shifts along the temporal dimension to align tokens at interleaved spatial positions. (2) The subsequent Multi-scale Channel Mixing (MCM) processes window-shifted tokens within a local spatial window at multiple time scales to mix dynamic information actively. After \verb|MHSA|, (3) the padding operation is employed ahead of (4) short-cutting, seamlessly assimilating dynamics into the encoding stream. 

\noindent {\bf{Window Shift Masking.}}
Given the substantial temporal redundancy inherent in videos, we propose to perform a tailored masking strategy.
Diverging from MAR~\cite{qing2023mar} which masks the frame patches for reconstruction, our WSM is employed to obtain interleaved spatial tokens for information interaction within a local spatial window and multiple temporal scales.
Formally, the spatial \textit{window} (\eg $w_1\times w_2 = 2\times 2$) is defined as the repeated unit used to partition all patches within each frame and generate masks. As shown in~\figref{pipeline}(b), the masking positions shift along the temporal dimension, sequentially discarding a specified proportion (\eg $r=0.5$) of tokens at different spatial locations.
The masking flow can be formulated as:
\begin{equation}
    M_{t}^l = \phi(M_{t-1}^l|M_{1:t-2}^l),l\in\{1,...,L\},
\end{equation}
where $M_{t}^l$ denotes the masking map for the $t$-th frame in the $l$-th ViT block and $\phi(\cdot|\cdot)$ is the periodic function to produce the masking map according to the former 1 to $(t-1)$ masking frames.
Then, the masking map $M_{t}^l$ is applied to visual tokens except for $\mathbf{z}_{t,cls}^{l-1}$ to yield window-shifted tokens:
\begin{equation}
    \mathbf{\dot{z}}^{l-1}_{t} = \mathbf{\dot{z}}^{l-1}_{t,1:N'} \gets M_{t}^{l}( \mathbf{z}_{t,1:N}^{l-1}),
\end{equation}
% \begin{equation}
%     k^{L-R+i}_{1:N',t} \gets k_{1:N,t}^{L-R+i}[M_{t,i}] = (W_k^{L-R+i} v^{L-R+i}_{1:N,t})[M_{t,i}]
% \end{equation}
where $\mathbf{\dot{z}}^{l-1}_{t}\in \mathbb{R}^{N'\times D}$ represent $N'(=r\times N)$ window-shifted spatial tokens of the $t$-th frame, and $N'=(\frac{L_1}{w_1}\times \frac{L_2}{w_2}) \times (r\times w_1\times w_2), L_1=\frac{H}{P}, L_2=\frac{W}{P}$. 
The WSM maintains essential spatiotemporal correlations for effective mixing of channel information across windows and time scales with MCM.

\noindent {\bf{Multi-scale Channel Mixing.}}
% Despite the aforementioned CLIP baseline considering multiple frame-level features through temporal averaging, a form of permutation-invariant space-time aggregation, hence, it indeed disregards the fundamental inter-frame dynamic signals in videos. 
After WSM, we follow the ``shift trick''~\cite{bulat2021space,wu2018shift} with zero-cost dimensionality reduction to perform MCM to enhance the fundamental inter-frame dynamic perception. 
% by capturing temporal coherence among multiple sets of frame tuples corresponding to different time intervals. 

Formally, to expand the dynamic insight of \verb|MHSA|, window-shifted tokens $\mathbf{\dot{z}}_t$ participate in fusing temporal information by indexing channels before and after the current time step. Here, we omit the denotation of layer index for simplicity. \figref{pipeline}(c) presents an example arranged in the window view to better illustrate the mechanism. When $\delta=2$, the visual tokens $\mathbf{\dot{z}}_{t}$ marked as ``1'' within each window absorb channel information from tokens $\mathbf{\dot{z}}_{t-\delta}$ and $\mathbf{\dot{z}}_{t+\delta}$ at ``3'', which formulates a local spatial window for interaction.
With the effect of temporal scales, the channel information flows actively among window-shifted tokens to capture spatiotemporal dynamics.
Let $\mathbf{\dot{z}}_t(d_s:d_e)\in \mathbb{R}^{N'\times(d_e-d_s)}$ be the operator for indexing the channels from $d_s$ to $d_e$ with $\mathbf{\dot{z}}_t$. For separate mixing, $\mathbf{\dot{z}}_t$ only interacts with $\mathbf{\dot{z}}_{t-\delta}$ and $\mathbf{\dot{z}}_{t+\delta}$ as:
\begin{equation}
    \mathbf{\dot{z}}_{t(\delta)} \triangleq [\mathbf{\dot{z}}_{t-\delta }(0:d_{\delta});\mathbf{\dot{z}}_{t+\delta}(d_{\delta}:2d_{\delta});\mathbf{\dot{z}}_{t}(2d_{\delta}:D)] \in \mathbb{R}^{N'\times D},
\end{equation}
where $d_\delta=\gamma\cdot D$ is a hyper-parameter for indexing channels. 
In contrast, for continual mixing (\eg $\delta=2$), $\mathbf{\dot{z}}_{t}$ is mixed with all window-shifted tokens of $2\delta$ range, \ie from $\mathbf{\dot{z}}_{t-2}$ to $\mathbf{\dot{z}}_{t+2}$:
\begin{equation}
\label{eq:scales}
\begin{aligned}
    \mathbf{\dot{z}}_{t(\delta)} \triangleq [\mathbf{\dot{z}}_{t-2 }(0:\frac{d_{\delta}}{\delta});\mathbf{\dot{z}}_{t-1 }(\frac{d_{\delta}}{\delta}:d_{\delta});\mathbf{\dot{z}}_{t+1}(d_{\delta}:d_{\delta}+\frac{d_{\delta}}{\delta});\\
    \mathbf{\dot{z}}_{t+2}(d_{\delta}+\frac{d_{\delta}}{\delta}:2d_{\delta});\mathbf{\dot{z}}_{t}(2d_{\delta}:D)] \in \mathbb{R}^{N' \times D},
\end{aligned}
\end{equation}
where the channel dimension for mixing each token is divided by $\delta$ to achieve zero-cost dimensionality reduction.
% Meanwhile, we obtain the mixing value tokens $\mathbf{\tilde{v}}'_{t_{\delta}}$ using an analogous operation as $\mathbf{\tilde{k}}'_{t_{\delta}}$.
Then, we perform \verb|MHSA| with the mixed tokens as follows:
\begin{equation}
    \mathbf{\dot{z}}_{t(\delta)}^{l} = \verb|MHSA|(\verb|LN|(\mathbf{\dot{z}}_{t(\delta)}^{l-1}))+\mathbf{\dot{z}}_{t(\delta)}^{l-1} \in \mathbb{R}^{N'\times D}.
\end{equation}

In a similar way, we further introduce multiple time scales $\{\delta_i\}_{i=1}^S$, with the purpose of unveiling the abundant dynamic insight and enriching the spatiotemporal information fusion. 
Then, we employ average pooling on all time scales to obtain $\mathbf{\bar{z}}_{t}^{l}\in \mathbb{R}^{N'\times D}$.

\noindent {\bf{Spatial Padding and Short-cutting.}}
Note that the quantity of visual tokens is reduced to $N'$ due to the aforementioned WSM operation. Therefore, we employ a straightforward spatial padding strategy to obtain $\mathbf{\tilde{z}}_t^l\in \mathbb{R}^{(N+1)\times D}$, where tokens $\mathbf{z}'^l_t$ computed by Eq.\eqref{eq:z} are selected carefully to fill in with $\mathbf{\bar{z}}_{t}^{l}$ according to $M_t^l$. As shown in~\figref{pipeline}(d), it restores the original number and spatial positions of tokens for seamless short-cutting to fuse additional dynamics effortlessly.

\noindent {\bf{Computational Complexity.}}
Notably, the overall computational complexity of our Space-time Cross Attention is $O(TN^2+ST(N')^2)=O(TN^2)$, which is equal to that of spatial-only attention. In contrast, the complexity of AIM and full space-time attention is $O(TN^2+T^2N)$ and $O(T^2N^2)$, respectively.

\subsection{Spatiotemporal Text Augmentation}
\label{sec:text}
In addition to expand the dynamic perception via visual representation learning, we further propose spatiotemporal text augmentation to incorporate refined action knowledge for visual-semantic context alignment by prompting GPT-3.5~\cite{achiam2023gpt}. For clarity, we present a detailed text augmentation process of \verb|abseiling| in \figref{pipeline}(2). 

\noindent {\bf{Action Semantic Knowledge Graph.}}
Essentially, our primary objective is to construct an ASKG that conceptually disentangles action categories into static appearances and dynamic motions and their interrelations, as shown in~\figref{pipeline}(e).
Specifically, the ASKG abstracts action categories into a graph structure by representing the original actions, related static and dynamic concepts as graph nodes and their interrelations as edges. 
Therefore, it enables a more structured and interpretable representation of actions.
We use the following prompts: 
``\textit{Return the object entity list containing Top $K(5\le K\le 10)$  most relevant objects / sub-actions involved in action:} \{\verb|abseiling|\}'' for semantic concepts, and ``\textit{Find the proper predicate names that concisely describe the relationship between each object / sub-action pair chosen from the entity list}'' for semantic relations.  \footnote{For a detailed demonstration of the prompts we used and response examples, please refer to Appendix.\label{fn:SMprompts}}

\noindent {\bf{Extending Text Prompts with Refined Knowledge.}}
To incorporate the refined spatiotemporal semantic knowledge into our multi-modal pipeline, we propose to extend text prompts by parsing the ASKG to generate text prompts, as presented in~\figref{pipeline}(f). 
Similar to the ASKG construction, we use the following prompts to extend the text prompts: ``\textit{Try to complete the whole sentence according to each relation triples: This is an example of} \{\verb|abseiling|\} \textit{,...}''\footref{fn:SMprompts}. In this way, the extended text prompts are then obtained by concatenating the hard prompt templates with the output. 
The generated clauses reflect different spatiotemporal text hints, maintaining semantic consistency in describing the action. 
% 
% We generate abundant text prompts according to the structural semantic information of the graph by extracting different object and action concepts, respectively. 
The spatial text prompts $C^s$ describe the static appearances obtained by prompting \textit{object relation triples}, while the temporal text prompts $C^t$ capture the dynamic motions by prompting \textit{action relation triples}. 

\subsection{Training Objectives}
\label{sec:FADA}
To overcome visual-semantic discrepancies at instance and frame level, the primary training objective is to meticulously align multi-modal spatiotemporal dynamics based on fine-grained alignment scores. 

Specifically, let $\mathbf{z}_{n,t}$ be the final visual representation of the $n$-th video at the $t$-th frame, the alignment score for the $k$-th class is calculated across $N^{st}$ spatiotemporal text representations $\{\mathbf{c}^{st}_{k,j}\}_{j=1}^{N^{st}}$ to achieve frame-to-prompt and symmetric prompt-to-frame fine-grained alignment:
\begin{equation}
\label{eq:score}
S^{v2t}_{n,k}= \frac{1}{T} \sum_{t=1}^{T} \max_{1 \le j \le N^{st}}  \mathbf{z}_{n,t}^\top \mathbf{c}^{st}_{k,j};\quad
S^{t2v}_{n,k}= \frac{1}{N^{st}} \sum_{j=1}^{N^{st}} \max_{1\le t \le T}  \mathbf{z}_{n,t}^\top \mathbf{c}^{st}_{k,j}.
\end{equation}

The overall alignment score $S_{n,k}$ is calculated by averaging the scores above, which is also used for our zero-shot inference.
The cross-entropy loss is implemented following~\cite{wu2023revisiting,jia2024generating} as:
\begin{equation}
    L_{CE} = -\frac{1}{B}\sum_{n=1}^{B} \sum_{k=1}^{K}y_{n,k}\log{\left ( \frac{\exp(S_{n,k})}{\sum_{i=1}^{K} \exp(S_{n,i})}  \right ) }, 
\end{equation}
where $B$ denotes the number of minibatch training videos of $K$ seen classes. If the $n$-th video belongs to the $k$-th class, $y_{n,k}$ equals 1; otherwise, $y_{n,k}$ equals 0. 
Finally, our framework is optimized by $L_{CE}$ together with feature distillation loss proposed by~\cite{huang2024froster}.

%%%%%%%%% Table1  %%%%%%%%%%%%%%%%
\begin{table*}[t]
\centering
\setlength{\tabcolsep}{1mm}
\label{ep1&2}
\begin{threeparttable}
\begin{tabular}{lccccccc}
\toprule
\multirow{2}{*}{Method} & \multirow{2}{*}{Venue} & \multirow{2}{*}{Encoder} & \multirow{2}{*}{SA Method\tnote{1}} & \multicolumn{2}{c}{UCF} & \multicolumn{2}{c}{HMDB} \\ \cmidrule{5-8} 
 &  &  &  & EP 1 & EP 2 & EP 1 & EP 2 \\ \hline
 \rowcolor{gray!20}\multicolumn{8}{l}{\textit{Uni-modal Vision Training}}\\ 
TS-GCN \cite{gao2019know} & AAAI'19 & GoogleNet & WE\tnote{1} & 36.14 ± 4.8 & - & 23.2 & - \\
CWEGAN \cite{mandal2019out} & CVPR'19 & I3D & WE & 26.9 ± 2.8 & - & 30.2 & - \\
% E2E \cite{brattoli2020rethinking} & CVPR'20 & R(2+1)D & WE & 48.0 & 32.7 & - & - \\
ER-ZSAR \cite{chen2021elaborative} & ICCV'21 & TSM & ED\tnote{1} & 51.8 ± 2.9 & - & 35.3 ± 4.6 & - \\
DASZL \cite{kim2021daszl} & AAAI'21 & TSM & HA\tnote{1} & 48.9 ± 5.8 & - & - & - \\ \hline
% REST \cite{lin2022cross} & CVPR'22 & ResNet101 & WE & 58.7 ± 3.3 & 40.6 & 41.1 ±3.7 & 34.4 \\ \hline
\rowcolor{gray!20}\multicolumn{8}{l}{\textit{Adapting Pretrained VLMs}}\\
BIKE \cite{wu2023bike} & CVPR'23 & ViT-L/14 & AT\tnote{1} & 86.6 ± 3.4 & 80.8 & 61.4 ± 3.6 & 52.8 \\ 

%  &  & ViT-B/16 & & 89.2 ± 2.1 & 83.7 & 60.9 ± 3.9 & 51.0 \\ 
% \multirow{-2}{*}{OTI~\cite{zhu2023orthogonal}} & \multirow{-2}{*}{ACM MM'23} & ViT-L/14  & \multirow{-2}{*}{CT\tnote{1}} & 92.8 ± 2.5  & \underline{88.3} & 64.0 ± 6.1 & 55.8  \\
&  &  & CT & 85.8 ± 3.3 & 79.6 & 58.1 ± 5.7 & 49.8 \\ 
\multirow{-2}{*}{Text4Vis*~\cite{wu2023revisiting}} & \multirow{-2}{*}{AAAI'23} & \multirow{-2}{*}{ViT-L/14}  & \cellcolor{gray!10}ST Aug. & \cellcolor{gray!10}89.5 ± 2.9  & \cellcolor{gray!10}84.2 & \cellcolor{gray!10}63.7 ± 3.2 &  \cellcolor{gray!10}52.9 \\

 &  & ViT-B/16 & & 89.9 ± 1.7 & 83.5 & 64.5 ± 4.5 & 53.2 \\ 
\multirow{-2}{*}{Open-VCLIP~\cite{weng2023open}} & \multirow{-2}{*}{ICML'23} & ViT-L/14  & \multirow{-2}{*}{CT} & \underline{93.1 ± 1.9}  & \underline{87.9} & \underline{68.5 ± 4.0} & \underline{58.3}  \\

ViLT-CLIP~\cite{wang2024vilt} & AAAI'24 & ViT-B/16 & PE\tnote{1} & - & 73.9 & - & 45.3 \\ 

\rowcolor{orange!20} 
\cellcolor{orange!20} & \cellcolor{orange!20} & ViT-B/16 & \cellcolor{orange!20} & 90.3 ± 1.7 & 85.3 & 64.7 ± 3.8 & 54.7 \\
\rowcolor{orange!20} 
\multirow{-2}{*}{\cellcolor{orange!20}Ours} & \multirow{-2}{*}{\cellcolor{orange!20}} & ViT-L/14 & \multirow{-2}{*}{\cellcolor{orange!20}ST Aug.} & \textbf{93.4 ± 2.2} & \textbf{88.6} & \textbf{68.7 ± 4.5} & \textbf{58.7} \\ \bottomrule

\end{tabular}
  \begin{tablenotes}
           \small{ \item[1] Semantic augmentation (SA); Word embeddings (WE); Elaborative descriptions (ED); Hand-craft attributes (HA); Category text prompts (CT); Attribute text prompts (AT); PE: Prompt embeddings.}
  \end{tablenotes}
\end{threeparttable}
\caption{Comparisons of ZSAR accuracies (\%) on UCF and HMDB with EP 1 and EP 2. ``*'' denotes our re-evaluation.}
\end{table*}
\section{Experiment}
\label{sec:experiment}
\subsection{Implementation Details}
% \noindent {\bf Datasets.}
We use Kinetics-400 (K400)~\cite{k400} dataset as the training set and evaluate our method under ZSAR settings on three popular benchmarks: UCF101 (UCF)~\cite{UCF101}, HMDB51 (HMBD)~\cite{HMDB51}, and Kinetics-600 (K600)~\cite{k600}, following the evaluation protocols: EP 1, EP 2 and EP 3 in~\cite{brattoli2020rethinking,ni2022expanding}.

We use the K400 pretrained models to directly perform cross-dataset ZSAR evaluation. Generally, we use two official CLIP backbones: ViT-B/16 and ViT-L/14. 
% The initial learning rate is $3.33\times 10^{-6}$ and is decayed to $3.33\times 10^{-8}$ utilizing the AdamW optimizer following a cosine decay scheduler. We warm up the training of models for 2 epochs and optimize the weights for another 20 epochs. 
In our proposed Space-time Cross Attention, we define the spatial size of a repeated window as $w_1\times w_2\ (w_1=w_2=2)$ with a mask ratio of 50\%, and specify the temporal scales to $[\pm1, \pm2]$.
Following~\cite{weng2023open}, the models learned in different epochs are averaged to improve generalizability.
Each video clip is uniformly sampled with 8 frames during training. For ZSAR evaluation, we use 3 temporal and 1 spatial views per video, and linearly aggregate the prediction results. 
% The experiments are conducted on 8 NVIDIA RTX 24G 4090 GPUs. 
More implementations are provided in Appendix. 

\subsection{Main Results}
We compare our method with state-of-the-art ZSAR methods on three benchmarks following commonly-used evaluation protocols. In Table 1, we categorize the previous methods based on the visual backbones and semantic augmentation methods, presenting a comprehensive comparative analysis on UCF and HMDB under EP 1 and EP 2. Note that there is a significant performance gap between the previous uni-modal vision training methods and the methods adapting pretrained VLMs to ZSAR. Among these, our method exhibits the best performance on UCF and HMDB when using either the ViT-B/16 or the ViT-L/14 encoder. In comparison with other VLM-based methods, our method achieves better performance than the second-best competitor, \ie Open-VCLIP, on both UCF and HMDB, with a margin up to 1.8\% and 1.5\% using the ViT-B/16 encoder. Remarkably, our framework with ViT-B/16 backbone even outperforms BIKE (ViT-L/14) by 7.8\% and 5.9\% on UCF and HMDB, respectively. When compared with Text4Vis, the performance gap is up to 9.0\% on UCF's EP 2. Furthermore, our semantic augmentation method has demonstrated strong adaptability. As a representative, by implementing our spatiotemporal text augmentation to Text4Vis, directly replacing its category prompts without any retraining, Text4Vis (w/ ST Aug.) has experienced promising improvements, with gains of $+$3.7\%, $+$4.6\%, $+$5.6\% and $+$3.1\%, respectively.
%%%%%%%%% Table2  %%%%%%%%%%%%%%%%
\begin{table*}[t]
\centering
\label{tab:ep3}
\begin{threeparttable}
\begin{tabular}{lccccc}
\toprule
% Method(cite), Veneue, Temporal Modeling SE Method, UCF, HMDB, K600
{Method} & {Venue}  & {SA Method} & {UCF} & {HMDB} & {K600} \\ \midrule
ActionCLIP \cite{wang2021actionclip} & arXiv'21 & CT & 58.3 ± 3.4 & 40.8 ± 5.4 & 66.7 ± 1.1 \\
X-CLIP \cite{ni2022expanding} & ECCV'22 & CT & 72.0 ± 2.3 & 44.6 ± 5.2 & 65.2 ± 0.4 \\
% A5 \cite{ju2022a5} & ECCV'22 & CT + PE\tnote{1} & 69.3 ± 4.2 & 44.3 ± 2.2 & 55.8 ± 0.7 \\
% ViFi-CLIP \cite{rasheed2023fine} & CVPR'23 & CT & 76.8 ± 0.7 & 51.3 ± 0.6 & 71.2 ± 1.0 \\
% Vita-CLIP \cite{wasim2023vita} & CVPR'23 & CT + PE\tnote{1} & 75.0 ± 0.6 & 48.6 ± 0.6 & 67.4 ± 0.5 \\
MAXI \cite{lin2023match} & ICCV'23 & VC\tnote{1} & 78.2 ± 0.7 & 52.3 ± 0.6 & 71.5 ± 0.8 \\
% MAP \cite{wang2023seeing} & ACM MM'23 & CT + PE & 76.4 ± 2.5 & 50.1 ± 5.4 & - \\
VicTR \cite{kahatapitiya2024victr} & CVPR'24 & CT + AT & 72.4 ± 0.3 & 51.0 ± 1.3 & - \\
OST \cite{chen2024ost} & CVPR'24 & STD\tnote{1} & 77.9 ± 1.3 & 54.9 ± 1.1 & 73.9 ± 0.8 \\
AP-CLIP \cite{jia2024generating} & ACM MM'24 & AP\tnote{1} & 82.4 ± 0.5 & 55.4 ± 0.8 & 73.4 ± 1.0 \\
Open-VCLIP++~\cite{wu2024building} & TPAMI'24 & VC & 83.9 ± 0.6 & \underline{55.6 ± 1.4} & 73.4 ± 0.8 \\
FROSTER~\cite{huang2024froster} & ICLR'24 & AD\tnote{1} & \underline{84.8 ± 1.1} & 54.8 ± 1.3 & \underline{74.8 ± 0.9} \\
\rowcolor{orange!20}
Ours &  & ST Aug. & \textbf{85.2 ± 1.2} & \textbf{55.9 ± 0.2} & \textbf{75.1 ± 0.7} \\ \bottomrule
\end{tabular}
\begin{tablenotes}
    \small{\item[1] VC: Visual captions; AP: Action-conditioned prompts; STD: Spatiotemporal descriptors; AD: Action descriptions}
\end{tablenotes}
\end{threeparttable}
\caption{Comparison with recent CLIP-based state-of-the-art on UCF, HMDB (EP 3) and K600 dataset. All methods are based on CLIP ViT-B/16. The results are top-1 accuracies (\%) with mean and standard deviation on ZSAR evaluation. }
\end{table*}
\begin{table}[t]
    \centering
    \setlength{\tabcolsep}{1mm}
    \begin{tabular}{lccc}
    \toprule
\textbf{SE Method} & \textbf{UCF} & \textbf{HMDB} & \textbf{K600} \\ \cmidrule{1-4}
CT~\cite{wang2021actionclip} & 83.8 & 54.0 & 73.5 \\
STD~\cite{chen2024ost} & 84.9 & 55.1 & 75.0 \\
\textbf{ST Aug. (Ours)} & \textbf{85.2} &\textbf{55.9} & \textbf{75.1} \\
    \bottomrule
    \end{tabular}
    \label{tab:ep2}
    \caption{Performance comparison (Top-1 Acc. (\%)) with different semantic augmentation methods. }
\end{table}

Table 2 shows the comparison results under EP 3 and K600 benchmark with different CLIP-based methods using the ViT-B/16 backbone. Despite most of the methods use category text prompts (CT) for semantic embeddings, recent methods based on text augmentation, including MAXI and Open-VCLIP++ with visual captions, AP-CLIP with action-conditional prompts, and FROSTER with action descriptions, yield consistent performance improvements. 
Compared to OST with STD, our ST Aug. not only describes spatial appearances and temporal evolutions of actions but also possess the capability to discover spatial and temporal interrelations with ASKG for nuanced text prompts, which surpasses the best previous knowledge-based OST by 7.3\% on UCF and AP-CLIP by 1.7\% on K600.
We also compare our proposed text augmentation with CT and STD based on our vision backbone for fair comparison in Table 3. 
It can be observed that our ST Aug. consistently surpasses STD and CT on three datasets. Besides, integrating with our backbone rather than OST, STD leads to a notable increase in performance with gains of 7.0\% and 1.1\% in UCF and K600, further demonstrating the superiority of our method's effective spatiotemporal dynamics modeling.

\begin{table}[t]
\centering

\label{tab:ab1}
\begin{tabular}{lccl}
\toprule
\textbf{Adaptation}                   & \textbf{WSM} & \textbf{MCM} & \textbf{UCF} \\ \cmidrule{1-4}
\multirow{1}{*}{Frozen}                    & \ding{55}   & \ding{55}   & 74.2   \\ \cmidrule{1-4}
\multirow{4}{*}{Prompt-based} & \ding{55} & Separate & 79.8 (\textcolor{blue}{+5.6})  \\
&  \ding{55}   &  Continual   &  80.1 (\textcolor{blue}{+5.9})  \\
&  \ding{51}  &   Separate  & 82.3 (\textcolor{blue}{+8.1})   \\ 
& \ding{51} & Continual & 82.7 (\textcolor{blue}{+8.5}) \\  \cmidrule{1-4}
\multirow{4}{*}{Fully-tuned} &  \ding{55}   &  Separate   &  84.5 (\textcolor{blue}{+10.3})   \\
&  \ding{55}   &   Continual  & 84.6 (\textcolor{blue}{+10.4}) \\
&  \ding{51}   &   Separate  & 85.0 (\textcolor{blue}{+10.8}) \\
&  \ding{51}   &   Continual  & 85.2 (\textcolor{blue}{+11.0}) \\
\bottomrule
\end{tabular}
\caption{Effect of different operations in Space-time Cross Attention and adaptation methods.
}
\end{table}

\begin{figure}[t]
    \centering
\includegraphics[width=\columnwidth]{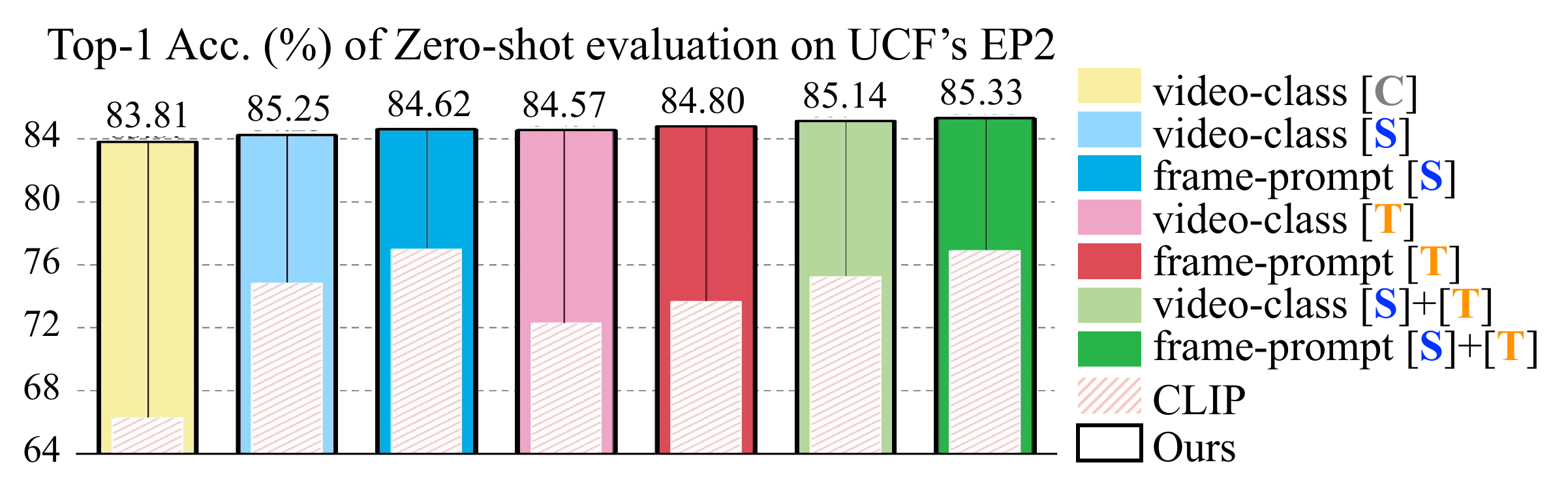}
    \caption{Effect of different combinations of text augmentation and alignment mechanisms for our method and CLIP.}
    \label{fig:ab3} 
\end{figure}
\begin{figure}[t]
    \centering
    \includegraphics[width=\columnwidth]{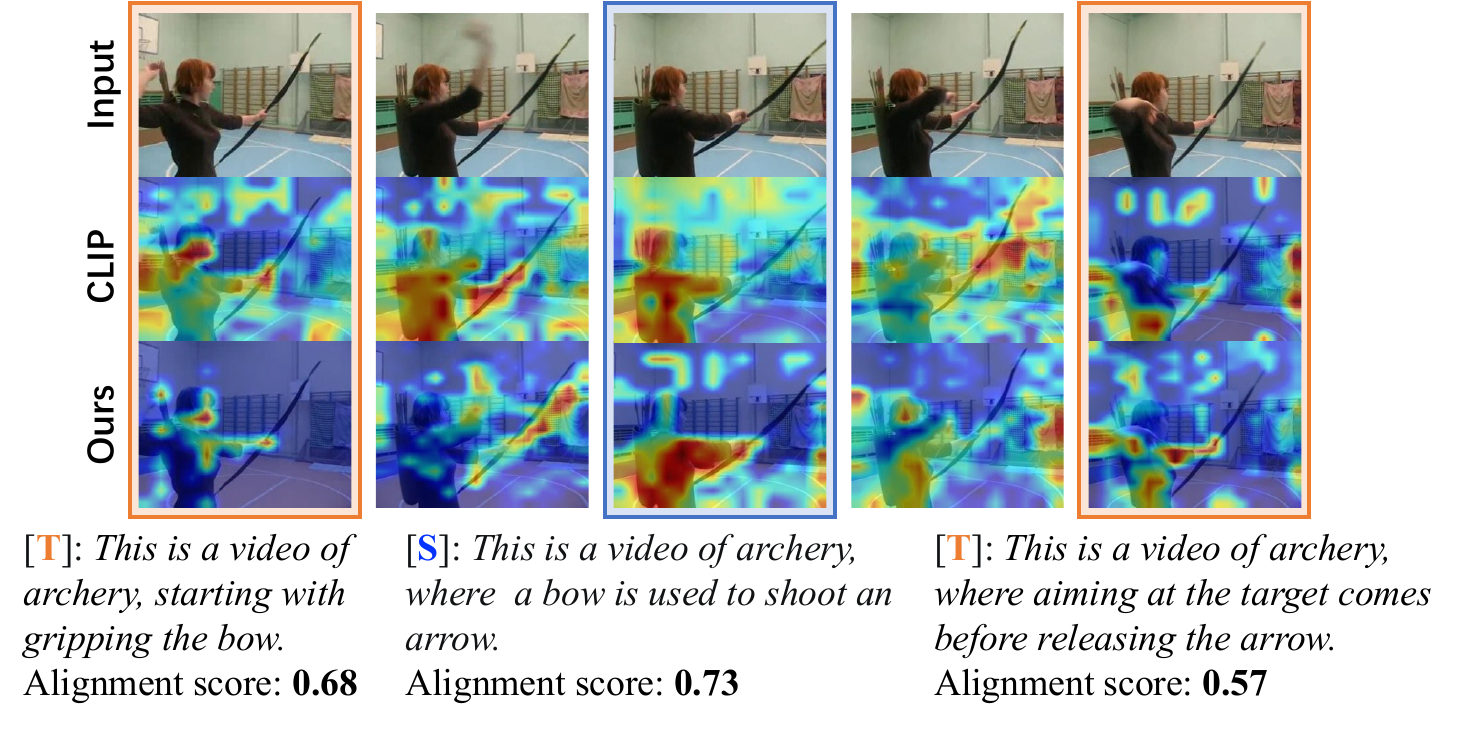}
    \caption{Visualizations of the attention maps and frame-prompt alignment scores of \textit{Archery}. Our framework consistently prioritizes local body parts and objects participated in the dynamic movements.}
    \label{fig:visual} 
\end{figure}
\subsection{Ablation Study}
\label{sec:ablation}
% To evaluate the impact of each component with our method, we conduct ablation studies using the ViT-B/16 backbone. 

\noindent {\bf Effect of Space-time Cross Attention.} Table 4 shows the effects of the Window Shift Masking (WSM) and Multi-scale Channel Mixing (MCM) with different adaptation methods. For a fair comparison, we start with the frozen CLIP as the baseline. In addition to our fully-tuned adaptation, we also evaluate prompt-based adaptation with the frozen backbone, where only a MLP is tuned to transform tokens after Space-time Cross Attention into prompts prepended with visual tokens within each block. However, fully-tuned adaptation consistently outperforms prompt-based adaptation, suggesting the importance of 
sufficient model capability for adapting to the video domain and the need for an effective space-time fusing strategy.
% enough capacity for models to adapt to video domain and the suitable space-time fusing strategy.
Additionally, by simply implementing MCM with fully-tuned adaptation, the accuracy on UCF is improved by 10.3\% and 10.4\% with continual mixing by capturing dynamics densely, achieving a slightly better performance than separate mixing.
% Additionally, integrating with WSM leads to a notable increase in performance, with separate mixing and continual mixing achieving gains of +5.6\% and +5.9\%, respectively. 
The most significant improvement is observed when WSM and MCM are used in conjunction, culminating in 85.0\%($+$10.8\%) and 85.2\%($+$11.0\%) on UCF.

\noindent {\bf Effect of spatiotemporal text augmentation and alignment mechanisms.} 
\figref{ab3} illustrates the effect of different implementations for text augmentation and alignment mechanisms. Notably, the proposed spatiotemporal text augmentation [{\textbf{S}}]+[{\textbf{T}}] and frame-prompt alignment significantly enhance performance of the vanilla CLIP on UCF without additional finetuning on the videos, highlighting its superior inference-time adaptability. However, for CLIP, the improvement attributable to spatiotemporal text prompts [{\textbf{S}}]+[{\textbf{T}}] is negligible compared with spatial text prompts [{\textbf{S}}], indicating potential object bias of the CLIP. 
In contrast, our proposed method achieves better results with spatiotemporal text prompts [{\textbf{S}}]+[{\textbf{T}}], showing a $+$1.08\% improvement over spatial text prompts [{\textbf{S}}], which validates its significant multi-modal spatiotemporal understanding capability. 

\subsection{Visualization}
\figref{visual} presents the attention map visualizations of video frames and text prompts with the highest alignment scores obtained by our method. 
The attention maps of CLIP primarily attend to the actor's body and surroundings unrelated to the actions being performed. Conversely, our framework consistently prioritizes the key body parts (\eg arms and hands) and objects (\eg bow and target) involved in the dynamic movements required for the action \textit{archery}, which is crucial for multi-modal spatiotemporal understanding. 

\section{Conclusion}
In this work, we present a novel multi-modal spatiotemporal expert for ZSAR. 
% Our intention centers on expanding the temporal perception of spatial attention by modeling temporal dynamics and refining category semantics. 
The Space-time Cross Attention integrates a four-step operation, capturing cross-frame dynamics without introducing additional parameters or increasing computational complexity.
The spatiotemporal text augmentation elaborates on static and dynamic concepts and their interrelations for action categories. 
Extensive evaluations consistently validate the superior effectiveness of our framework.

\section*{Acknowledgments}
This work is supported by National Natural Science Foundation of China (No. 62376217, 62301434), Young Elite Scientists Sponsorship Program by CAST (2023QNRC001), Key R\&D Project of Shaanxi Province (No. 2023-YBGY-240), and Young Talent Fund of Association for Science and Technology in Shaanxi, China (No. 20220117).
{\small
\bibliography{aaai25}
}
\newpage

\appendix
\section*{Supplementary Material}
This supplementary material offers extensive additional details and more experimental analysis complementing the main paper. The content is organized as follows:
\begin{enumerate}[label=\Alph*.]
    % \item Details of Multi-scale Temporal Prompt Module (\appenref{MTP})
    \item Prompts for LLMs and Responses (\appenref{prompts})
    \item Details of Datasets and Evaluation Protocols (\appenref{imple})
    \item Additional Ablation Studies (\appenref{results})
    \item Broader Impacts (\appenref{impacts})
    \item Limitations (\appenref{limitations})
\end{enumerate}

\section{Prompts for LLMs and Responses}
\label{appen:prompts}
\subsection{Details of Prompts for LLMs}
\begin{figure*}[p]
    \centering
    \includegraphics[width=\textwidth]{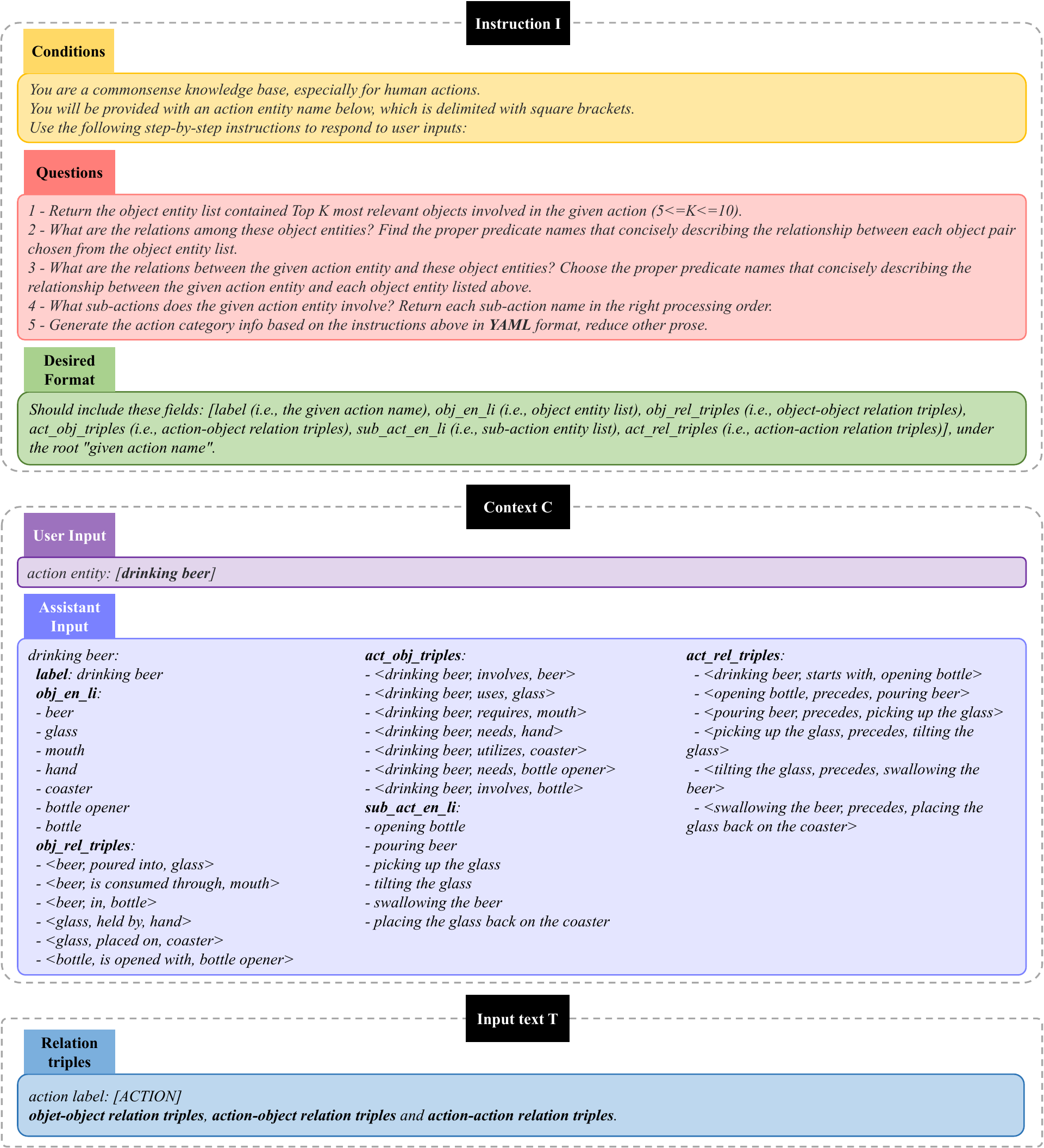}
    \caption{Prompts for Action Semantic Knowledge Graph (ASKG) construction. \textit{Instruction} \textbf{\textit{I}} specifies the conditions and required questions for entity list and relation triples, and constrains the desired format for the output. \textit{Context} \textbf{\textit{C}} provides an example input for the role of user and assistant. By inputting each category name in \textit{Input Text}, \textbf{\textit{T}}, we can then obtain the generated sub-graph of each seen and unseen class.} 
    \label{fig:ASKG} 
\end{figure*}

\begin{figure*}[p]
    \centering
    \includegraphics[width=\textwidth]{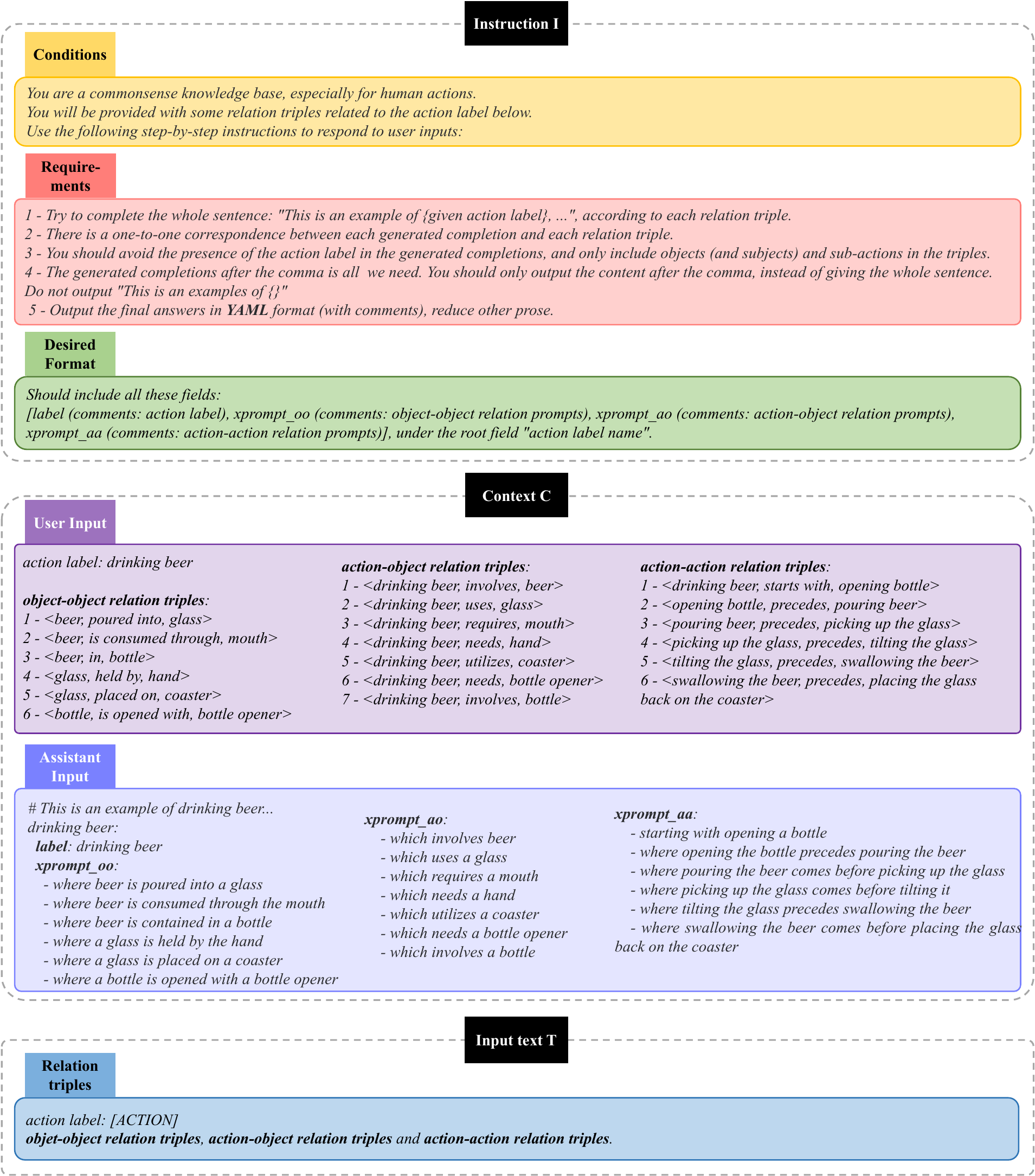}
    \caption{Prompts for Spatiotemporal Text Augmentation. \textit{Instruction} \textbf{\textit{I}} specifies the conditions and requirements for sentence completion, and constrains the desired format for the output. \textit{Context} \textbf{\textit{C}} presents an example input for the role of user and assistant. By inputting relation triples in \textit{Input Text}, \textbf{\textit{T}}, we can then obtain the augmented spatial and temporal text prompts corresponding to each triple.}
    \label{fig:STAug} 
    
\end{figure*}

In this section, we provide our complete prompts for ASKG construction and spatiotemporal text augmentation in a two-stage prompting manner in~\figref{ASKG} and~\figref{STAug}, respectively. Specially, we design the prompts with the idea of Chain-of-thought (CoT), which comprises three main elements, \ie \textit{Instruction} \textbf{\textit{I}}, \textit{Context} \textbf{\textit{C}} and \textit{Input text} \textbf{\textit{T}}. 
%%% instruction, context and input for ASKG and SSTD
%%%%%
\textit{Instruction} is a stated sentence that conducts the model to perform a specific task. Here, we first restrict the specified conditions of the LLM (\ie GPT-3.5-turbo) to serve as a ``\textit{commonsense knowledge base for human actions}'' to obtain a more consistent response to the requirements of the answer.
Next, with different purposes, we design several questions and express the requirements for these two prompting tasks. Note that the \textit{YAML} format output is preferred to obtain more efficient prompting results and to reduce token consumption.
As the second part of our prompts, \textit{Context} provides contextual knowledge of the input text and one-shot prompt instances consist of user input and assistant input, which mainly constrains the specific and consistent format and content of the output.
For the ASKG construction in~\figref{ASKG}, we concatenate each element in \textbf{\textit{T}} with \textbf{\textit{I}} and \textbf{\textit{C}} to get responses of each categories.
For spatiotemporal text augmentation shown in~\figref{STAug}, with ASKG accessed by the LLM, different spatiotemporal semantic information in ASKG is extracted, and LLM then uses these relation triples to generate corresponding descriptions from sentence forms (or non-predicate verb forms) rather than complete descriptions. Finally, the hard prompt templates are spliced with the generated clauses to obtain different spatiotemporal text hints, and the main sentence and clause maintain semantic consistency in describing the action. 


\subsection{Examples of Spatiotemporal Text Prompts Generation with Responses from LLMs}
\label{appen:response}
\noindent\textbf{Responses for action category ``\texttt{Archery}'':}

{\noindent\textit{Object list for \textbf{spatial text prompts}} [\textcolor{blue}{\textbf{S}}]:}
    {\begin{enumerate}
        \small
        \setlength{\itemsep}{0pt}%
        \item \texttt{bow}: \textit{This is a video of archery, which requires a bow.}
        \item \texttt{arrow}: \textit{This is a video of archery, which uses an arrow.}
        \item \texttt{target}: \textit{This is a video of archery, which aims at a target.}
        \item \texttt{quiver}: \textit{This is a video of archery, which utilizes a quiver.}
        \item \texttt{armguard}: \textit{This is a video of archery, which requires an armguard.}
        \item \texttt{finger tab}: \textit{This is a video of archery, which needs a finger tab.}
        \item \texttt{bullseye}: \textit{This is a video of archery, which involves a bullseye.}
    \end{enumerate}}

{\small\noindent\textit{Sub-action list:}
    {\begin{enumerate}
        \small
        \setlength{\itemsep}{0pt}%
        \item \texttt{gripping the bow}
        \item \texttt{nocking the arrow}
        \item \texttt{drawing the bowstring}
        \item \texttt{aiming at the target}
        \item \texttt{releasing the arrow}
        \item \texttt{following through}
    \end{enumerate}}

{\small\noindent\textit{Object relation triples for \textbf{spatial text prompts}} [\textcolor{blue}{\textbf{S}}]:}
    {\begin{enumerate}
        \small
        \setlength{\itemsep}{0pt}%
        \item \texttt{<bow, used to shoot, arrow>}: \textit{This is a video of archery, where a bow is used to shoot an arrow.}
        \item \texttt{<arrow, aimed at, target>}: \textit{This is a video of archery, where an arrow is aimed at a target.}
        \item \texttt{<quiver, holds, arrows>}: \textit{This is a video of archery, where arrows are held in a quiver.}
        \item \texttt{<armguard, protects, arm>}: \textit{This is a video of archery, where an armguard protects the arm.}
        \item \texttt{<finger tab, protects, fingers>}: \textit{This is a video of archery, where a finger tab protects the fingers.}
        \item \texttt{<target, has, bullseye>}: \textit{This is a video of archery, where a target has a bullseye.}
    \end{enumerate}}

{\small\noindent\textit{Sub-action relation triples for \textbf{temporal text prompts}} [\textcolor{orange}{\textbf{T}}]:}
    {\begin{enumerate}
        \small
        \setlength{\itemsep}{0pt}%
        \item \texttt{<archery, starts with, gripping the bow>}: \textit{This is a video of archery, starting with gripping the bow.}
        \item \texttt{<gripping the bow, precedes, nocking the arrow>}: \textit{This is a video of archery, where gripping the bow precedes nocking the arrow.}
        \item \texttt{<nocking the arrow, precedes, drawing the bowstring>}: \textit{This is a video of archery, where nocking the arrow comes before drawing the bowstring.}
        \item \texttt{<drawing the bowstring, precedes, aiming at the target>}: \textit{This is a video of archery, where drawing the bowstring precedes aiming at the target.}
        \item \texttt{<aiming at the target, precedes, releasing the arrow>}: \textit{This is a video of archery, where aiming at the target comes before releasing the arrow.}
        \item \texttt{<releasing the arrow, precedes, following through>}: \textit{This is a video of archery, where releasing the arrow precedes following through.}
    \end{enumerate}}

\noindent\textbf{Responses for action category ``\texttt{Surfing}'':}

{\noindent\textit{Object list for \textbf{spatial text prompts}} [\textcolor{blue}{\textbf{S}}]:}
    {\begin{enumerate}
        \small
        \setlength{\itemsep}{0pt}%
        \item \texttt{surfboard}: \textit{This is a video of surfing, which requires a surfboard.}
        \item \texttt{waves}: \textit{This is a video of surfing, which takes place on waves.}
        \item \texttt{ocean}: \textit{This is a video of surfing, which takes place in the ocean.}
        \item \texttt{beach}: \textit{This is a video of surfing, which takes place near a beach.}
        \item \texttt{wetsuit}: \textit{This is a video of surfing, which requires a wetsuit.}
        \item \texttt{leash}: \textit{This is a video of surfing, which uses a leash.}
        \item \texttt{fins}: \textit{This is a video of surfing, which uses fins.}
    \end{enumerate}}

{\small\noindent\textit{Sub-action list}:}
\texttt{
    {\begin{enumerate}
        \small
        \setlength{\itemsep}{0pt}%
        \item paddling out
        \item catching a wave
        \item standing up on the surfboard
        \item riding the wave
        \item performing tricks
        \item exiting the wave
    \end{enumerate}}
}

{\small\noindent\textit{Object relation triples for \textbf{spatial text prompts}} [\textcolor{blue}{\textbf{S}}]:}
    {\begin{enumerate}
        \small
        \setlength{\itemsep}{0pt}%
        \item \texttt{<surfboard, used on, waves>}: \textit{This is a video of surfing, using a surfboard on waves.}
        
        \item \texttt{<waves, found in, ocean>}: \textit{This is a video of surfing, finding waves in the ocean.}
        \item \texttt{<ocean, located near, beach>}: \textit{This is a video of surfing, the ocean being located near a beach.}
        \item \texttt{<wetsuit, worn during, surfing>}: \textit{This is a video of surfing, wearing a wetsuit during surfing.}
        \item \texttt{<leash, attached to, surfboard>}: \textit{This is a video of surfing, attaching a leash to the surfboard.}
        \item \texttt{<fins, attached to, surfboard>}: \textit{This is a video of surfing, attaching fins to the surfboard.}
    \end{enumerate}}

{\small\noindent\textit{Sub-action relation triples for \textbf{temporal text prompts}} [\textcolor{orange}{\textbf{T}}]:}
    {\begin{enumerate}
        \small
        \setlength{\itemsep}{0pt}%
        \item \texttt{<surfing, starts with, paddling out>}: \textit{This is a video of surfing, starting with paddling out.}
        \item \texttt{<paddling out, precedes, catching a wave>}: \textit{This is a video of surfing, where paddling out precedes catching a wave.}
        \item \texttt{<catching a wave, precedes, standing up on the surfboard>}: \textit{This is a video of surfing, attaching fins to the surfboard.}
        \item \texttt{<standing up on the surfboard, precedes, riding the wave>}: \textit{This is a video of surfing, where standing up on the surfboard precedes riding the wave.}
        \item \texttt{<riding the wave, precedes, performing tricks>}: \textit{This is a video of surfing, where riding the wave comes before performing tricks.}
        \item \texttt{<performing tricks, precedes, exiting the wave>}: \textit{This is a video of surfing, where performing tricks precedes exiting the wave.}
    \end{enumerate}}

\noindent\textbf{Responses for action category ``\texttt{Clean and jerk}'':}

{\noindent\textit{Object list for \textbf{spatial text prompts}} [\textcolor{blue}{\textbf{S}}]:}
    {\begin{enumerate}
        \small
        \setlength{\itemsep}{0pt}%
        \item \texttt{barbell}: \textit{This is a video of clean and jerk,  which involves a barbell.}
        \item \texttt{platform}: \textit{This is a video of clean and jerk,  which requires a platform.}
        \item \texttt{chalk}: \textit{This is a video of clean and jerk,  which uses chalk.}
        \item \texttt{weightlifting belt}: \textit{This is a video of clean and jerk,  which requires a weightlifting belt.}
        \item \texttt{weightlifting shoes}: \textit{This is a video of clean and jerk,  which requires weightlifting shoes.}
        \item \texttt{grip}: \textit{This is a video of clean and jerk, which involves grip.}
        \item \texttt{shoulders}: \textit{This is a video of clean and jerk, which utilizes the shoulders.}
        \item \texttt{legs}: \textit{This is a video of clean and jerk, which utilizes the legs.}
        \item \texttt{arms}: \textit{This is a video of clean and jerk, which utilizes the arms.}
    \end{enumerate}}

{\small\noindent\textit{Sub-action list}:}
\texttt{
    {\begin{enumerate}
        \small
        \setlength{\itemsep}{0pt}%
        \item setting up the barbell
        \item gripping the barbell
        \item cleaning the barbell
        \item jerking the barbell
    \end{enumerate}}
}

{\small\noindent\textit{Object relation triples for \textbf{spatial text prompts}} [\textcolor{blue}{\textbf{S}}]:}
    {\begin{enumerate}
        \small
        \setlength{\itemsep}{0pt}%
        \item \texttt{<barbell, loaded with, weights>}: \textit{This is a video of clean and jerk, where a barbell is loaded with weights.}
        \item \texttt{<barbell, placed on, platform>}: \textit{This is a video of clean and jerk, where a barbell is placed on a platform.}
        \item \texttt{<chalk, used for, grip>}: \textit{This is a video of clean and jerk, where chalk is used for grip.}
        \item \texttt{<weightlifting belt, worn on, waist>}: \textit{This is a video of clean and jerk, where a weightlifting belt is worn on the waist.}
        \item \texttt{<weightlifting shoes, worn on, feet>}: \textit{This is a video of clean and jerk, where weightlifting shoes are worn on the feet.}
        \item \texttt{<grip, on, barbell>}: \textit{This is a video of clean and jerk, where there is a grip on the barbell.}
        \item \texttt{<shoulders, support, barbell>}: \textit{This is a video of clean and jerk, where the shoulders support the barbell.}
        \item \texttt{<legs, provide power for, lift>}: \textit{This is a video of clean and jerk, where the legs provide power for the lift.}
        \item \texttt{<arms, lift, barbell>}: \textit{This is a video of clean and jerk, where the arms lift the barbell.}
    \end{enumerate}}

{\small\noindent\textit{Sub-action relation triples for \textbf{temporal text prompts}} [\textcolor{orange}{\textbf{T}}]:}
    {\begin{enumerate}
        \small
        \setlength{\itemsep}{0pt}%
        \item \texttt{<clean and jerk, starts with, setting up the barbell>}: \textit{This is a video of clean and jerk, starting with setting up the barbell.}
        \item \texttt{<setting up the barbell, precedes, gripping the barbell>}: \textit{This is a video of clean and jerk, where setting up the barbell precedes gripping the barbell.}
        \item \texttt{<gripping the barbell, precedes, cleaning the barbell>}: \textit{This is a video of clean and jerk, where gripping the barbell precedes cleaning the barbell.}
        \item \texttt{<cleaning the barbell, precedes, jerking the barbell>}: \textit{This is a video of clean and jerk, where cleaning the barbell precedes jerking the barbell.}
    \end{enumerate}}

\section{Details of Datasets and Evaluation Protocols}
\label{appen:imple}
\subsection{Datasets}
\label{appen:dataset}
We conduct the training process on Kinetics-400~\cite{k400} dataset and perform evaluations on other three popular benchmarks: UCF101~\cite{UCF101}, HMDB51~\cite{HMDB51}, and Kinetics-600~\cite{k600}. 

\noindent {\bf Kinectics-400 and Kinectics-600} are both comprehensive video datasets for human action recognition. Kinectics-400 contains 400 action categories of approximately 240k training and 20k validation videos collected from YouTube, which covers a wide range of human actions, including sports activities, daily life actions, and various interactions, serving as a widely-used action recognition dataset for pretraining. The duration of video clips in Kinetics-400 varies, with most clips being around 10 seconds long. This diversity in video duration helps models learn temporal dynamics and context for action recognition. Kinectics-600 extends Kinectics-400 by incorporating an additional 220 new categories, thus enabling to evaluate zero-shot learning capabilities on these novel categories. 

\noindent {\bf UCF-101} is a human action recognition dataset collected from YouTube, and consists of 13,320 video clips, which are classified into 101 categories. These 101 categories encompass a wide range of realistic actions including body motion, human-human interactions, human-object interactions, playing musical instruments and sports. Officially, there are three splits allocating 9,537 videos for training and 3,783 videos for testing.

\noindent {\bf HMDB-51} is a relatively small video dataset compared to the Kinetics and UCF-101, and is composed of 6,766 videos across 51 action categories (such as “jump”, “kiss” and “laugh”), ensuring at least 101 clips within each category. The original evaluation scheme employs three distinct training/testing splits, allocating 70 clips for training and 30 clips for testing of each category in each split.

\subsection{Evaluation Protocols}
\label{appen:ep}
For ZSAR evaluation on HMDB-51 and UCF-101, we follow~\cite{brattoli2020rethinking} as below:
\begin{itemize}
\item Evaluation Protocol 1 (EP 1): We randomly choose half of the test set classes as the selected subset for each dataset, i.e., 50 categories for UCF-101 and 25 categories for HMBD-51. Repeat ten times and report the averaged results for each subset.

\item Evaluation Protocol 2 (EP 2): We directly evaluate our method on the full 101 categories of UCF-101 and 51 categories of HMDB-51.

\item Evaluation Protocol 3 (EP 3): We use three official test splits of each dataset as ZSAR evaluation, and report the averaged results of these splits.
\end{itemize}

For ZSAR evaluation on Kinectics-600, we use the same splits as~\cite{ni2022expanding}, which randomly splits 220 new categories into 60 validation categories plus 160 test categories by repeating three times. The average top-1 accuracy and standard deviation are reported.

\section{Additional Ablation Studies}
\label{appen:results}
\begin{figure}[t]
    \centering
    \includegraphics[width=\columnwidth]{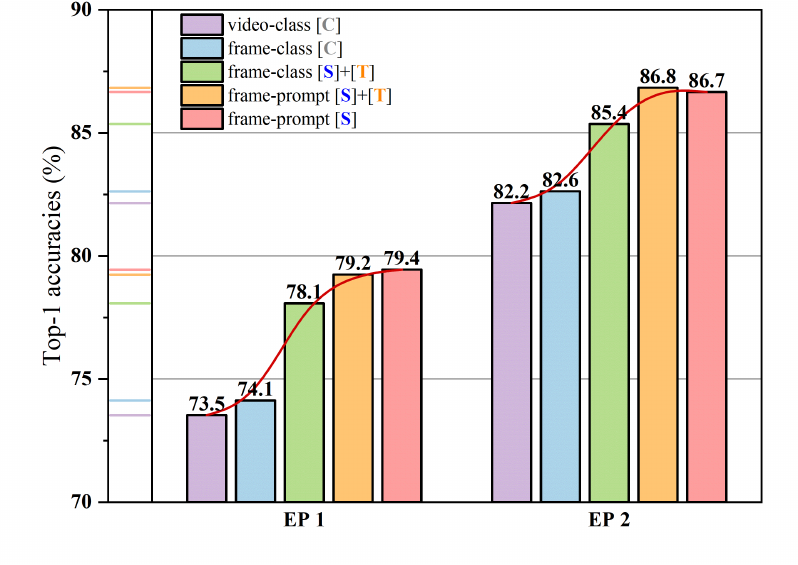}
    \caption{Effect of different combinations of text augmentation and alignment mechanisms for CLIP ViT-L/14.}
    \label{fig:add} 
\end{figure}
In this part, we provide additional ablation studies for the in-depth analysis of the effect of other complements in our proposed method, including different masking strategies and masking ratios in WSM, multiple temporal scales in MCM, text augmentation and fine-grained alignment for CLIP.

\begin{table}[t]
\centering
\label{tab:aab1}
\renewcommand{\arraystretch}{1.5}
\begin{tabular}{cccll}
\toprule
\multicolumn{1}{c}{$r (\%)$} & Spatial Mask &  Temp. Shift & UCF & HMDB \\ \hline
25 & Repeat Window & \ding{51} & 85.0 & 55.5 \\ \hline
\multirow{5}{*}{50} 
 & Random & \ding{55} & 79.6 & 49.6 \\
 & Random & \ding{51} & 83.7 & 53.1 \\
 & Uniform & \ding{51} & 84.8 & 54.9 \\
& Rand. Window & \ding{51} & 85.1 & 55.6 \\
 & \cellcolor{gray!20}Repeat Window & \cellcolor{gray!20}\ding{51} & \cellcolor{gray!20}\textbf{85.2} & \cellcolor{gray!20}\textbf{55.9} \\ \hline
75 & Repeat Window & \ding{51} & 85.0 & 55.7 \\
\bottomrule
\end{tabular}
\caption{Effect of different spatial and temporal shift masking strategies and mask ratios $r(\%)$ with the ViT/B-16 backbone under EP 3. The spatial window size is $2\times 2$ for both random window and repeat window spatial masking.}
\end{table}

\begin{table}[t]
    \centering
    \renewcommand{\arraystretch}{1.5}
    \begin{tabular}{ccc}
    \toprule
\multicolumn{1}{l}{Temporal Scales} & UCF (\%) & HMDB (\%) \\ \hline
$[\pm 1]$ & 84.9 & 55.7 \\
$[\pm 2]$ & 84.8 & 55.5 \\
$[\pm 3]$ & 84.1 & 55.3 \\

\cellcolor{gray!20}$[\pm 1, \pm 2]$ & \cellcolor{gray!20}85.2 & \cellcolor{gray!20}55.9 \\
$[\pm 1, \pm 2, \pm 3]$ & 85.3 & 55.6 \\
    \bottomrule
    \end{tabular}
    \caption{Effect of multiple temporal scales in MCM with the ViT-B/16 backbone under EP 3.}
    \label{tab:aab2}
\end{table}
\noindent {\bf Effect of different masking strategies and ratios in WSM.}
We compare the masking strategy in WSM (\ie repeat window + temporal shift) with other spatial mask sampling and temporal shift strategies under different mask ratios $r(\%)$ in Table 1. 
With $r=50\%$, we perform ``random spatial masking'' for each frame individually to leave out spatiotemporal correlations without temporal shift. 
Beside, we generate random and uniform masks for the first frame, and follow the temporal shift strategy to mask the subsequent frames. In contrast to our MSM, with ``random window spatial masking'', each window within the frame is masked randomly rather than consistently. The observations can be summarized from the table as: (1) The temporal shift masking strategy significantly boost the performance of random spatial masking (79.6\% \vs 83.7\%, 49.6\% \vs 55.9\%), which validates the importance of maintaining spatiotemporal correlations for video understating; 
(2) Compared with other spatial masking strategies, our ``repeat window spatial masking'' yields highest performance on both UCF and HMDB. We observe that the ``random window spatial masking'' results in a slight decrease in performance of $-0.1\%$ and $-0.3\%$. This implies that masking with consistency and reasonable randomness lead to greater improvements;
(3) By comparing our WSM under different masking ratios, it can be noticed that more gains can be achieved with $r=50\%$. This suggests that discarding half of the contexts can introduce a more effective temporal dynamic modeling and balanced spatiotemporal fusion.

\noindent {\bf Effect of multiple temporal scales in MCM.}
We investigate the effect of MCM when attending to different temporal scales in Table 2. It can be observed that the performance consistently decreases when expanding the single temporal scale from $[\pm 1]$ to $[\pm 3]$. This can be explained by the fact that larger temporal scales result in sparser interactions for boundary frames during channel mixing.
Better outcomes are observed when introducing multiple temporal scales of $[\pm 1, \pm 2]$ and $[\pm 1, \pm 2, \pm 3]$, where the combination of $[\pm 1, \pm 2]$ achieves better UCF-HMDB accuracy trade-off than other temporal scales. 

\noindent {\bf Different combinations of text augmentation and alignment mechanism for CLIP.}
For a fair comparison, we start with CLIP baselines that obtain video representations via average pooling and use category text prompts [\textcolor{gray}{\textbf{C}}] to perform video-class alignment. We then introduce different text prompts, including spatial [\textcolor{blue}{\textbf{S}}], temporal [\textcolor{orange}{\textbf{T}}] and spatiotemporal [\textcolor{blue}{\textbf{S}}]+[\textcolor{orange}{\textbf{T}}], with frame-prompt and video-class alignment that is obtained by average pooling to the video and category, to validate the effect of our spatiotemporal text augmentation. 
As shown in~\figref{add}, the proposed spatiotemporal text augmentation [\textcolor{blue}{\textbf{S}}]+[\textcolor{orange}{\textbf{T}}] and frame-prompt alignment significantly enhance performance of the vanilla CLIP on UCF-101 dataset without additional training on the video dataset, achieving top-1 accuracies of 79.2\% and 86.8\% with EP 1 and EP 2, respectively.
It is noteworthy that the common practice to obtain video-level representations by average pooling the frame-level representations results in a performance decrease of 0.6\% and 0.4\% with [\textcolor{gray}{\textbf{C}}] text prompts, and 1.1\% and 1.4\% with [\textcolor{blue}{\textbf{S}}]+[\textcolor{orange}{\textbf{T}}] text prompts, respectively. This implies that inappropriate feature aggregation may harm the discriminative frame-level representations obtained by CLIP.

\section{Broader Impacts}
\label{appen:impacts}
\noindent {\bf Positive Impacts.} 
1) Our work is the first to emphasize the importance of collaborative multi-modal spatiotemporal understanding for ZSAR, encouraging researchers to develop more comprehensive, balanced and fine-grained zero-shot learners. 
2) The proposed Space-time Cross Attention capture cross-frame dynamics without introducing additional parameters or increasing computational complexity. It has broad potential for application in other video understanding tasks with ViT-based encoders, enabling efficient and effective spatiotemporal modeling.
3) The proposed Action Semantic Knowledge Graph (ASKG) highlights contextual information about actions, including their temporal and spatial concepts, as well as their dependencies and associations with other actions. Due to the structural knowledge and flexibility of the ASKG, we can foresee its broader benefits on other tasks, such as action anticipation and reasoning, human-object interaction detection and so on. Furthermore, we also encourage the implementations of building a comprehensive ASKG with additional abnormal actions following the similar prompts to act as a cornerstone for promising intelligent video understanding systems.
4) The proposed \textbf{Spatiotemporal Dynamic Duo} framework showcases superior effectiveness under the challenging ZSAR settings. With the strong generalization capability, our framework can empower many downstream applications, such as human-computer interaction, autonomous vehicles, surveillance and security \etc.

\noindent {\bf Negative Impacts.}
1) The dependency on LLMs could perpetuate biases, inequalities, and misinformation inherent in the data utilized for pretraining the models.
2) Since surveillance and security could be applications of this technology, there are potential risks of privacy violations, data leakage, and other security concerns.

\section{Limitations}
\label{appen:limitations}
Despite achieving promising ZSAR performance with collaborative multi-modal spatiotemporal understanding, our method still suffers some limitations. 
Due to the hallucination of LLMs, the quality of spatiotemporal text prompts of different categories varies thus impacting the final performance. Moreover, purely text-based prompts, even if of satisfactory quality, may still not be ideal for enhancing action recognition performance due to the absence of visual cues during text augmentation. A multi-modal, knowledge-enhanced text augmentation approach may be more suitable for addressing this issue, and we will explore this further in our future work.

\end{document}